\documentclass[conference]{IEEEtran}
\IEEEoverridecommandlockouts
\usepackage{cite}
\usepackage{placeins}
\usepackage{amsmath,amssymb,amsfonts}
\usepackage{algorithmic}
\usepackage{graphicx}
\usepackage{textcomp}
\usepackage{xcolor}
\def\BibTeX{{\rm B\kern-.05em{\sc i\kern-.025em b}\kern-.08em
    T\kern-.1667em\lower.7ex\hbox{E}\kern-.125emX}}
\begin{document}

\title{Multi-Objective Optimization for Size and Resilience of Spiking Neural Networks 
\thanks{
This manuscript has been authored by UT-Battelle, LLC under Contract
No. DE-AC05-00OR22725 with the U.S. Department of Energy.  The United
States Government retains and the publisher, by accepting the article
for publication, acknowledges that the United States Government
retains a non-exclusive, paid-up, irrevocable, worldwide license to
publish or reproduce the published form of this manuscript, or allow
others to do so, for United States Government purposes. The
Department of Energy will provide public access to these results of
federally sponsored research in accordance with the DOE Public Access
Plan (http://energy.gov/downloads/doe-public-access-plan).}
}

\author{
    \IEEEauthorblockN{Mihaela Dimovska,\IEEEauthorrefmark{1} \quad Travis Johnston,\IEEEauthorrefmark{2} \quad Catherine D. Schuman,\IEEEauthorrefmark{2}}
     \IEEEauthorblockN{J. Parker Mitchell,\IEEEauthorrefmark{2} \quad
     Thomas E. Potok \IEEEauthorrefmark{2} }
    \IEEEauthorblockA{\IEEEauthorrefmark{1}\textit{Department of Electrical and Computer Engineering}, \\
\textit{University of Minnesota, 
Minneapolis, USA} \\
dimov003@umn.edu}
   \IEEEauthorblockA{\IEEEauthorrefmark{2}\textit{Oak Ridge National Laboratory, 
Oak Ridge, USA} \\
\{johnstonjt, schumancd, mitchelljp1, potokte\}@ornl.gov}
}

\IEEEoverridecommandlockouts
\IEEEpubid{\makebox[\columnwidth]{ 978-1-7281-3885-5/19/\$31.00~\copyright2019 IEEE \hfill} \hspace{\columnsep}\makebox[\columnwidth]{ }}

\maketitle

\IEEEpubidadjcol

\begin{abstract}
Inspired by the connectivity mechanisms in the brain, neuromorphic computing architectures model Spiking Neural Networks (SNNs) in silicon. As such, neuromorphic architectures are designed and developed with the goal of having small, low power chips that can perform control and machine learning tasks. However, the power consumption of the developed hardware can greatly depend on the size of the network that is being evaluated on the chip. Furthermore, the accuracy of a trained SNN that is evaluated on chip can change due to voltage and current variations in the hardware that perturb the learned weights of the network. While efforts are made on the hardware side to minimize those perturbations, a software based strategy to make the deployed networks more resilient can help further alleviate that issue. 
In this work, we study Spiking Neural Networks in two neuromorphic architecture implementations with the goal of decreasing their size, while at the same time increasing their resiliency to hardware faults. We leverage an evolutionary algorithm to train the SNNs and propose a multi-objective fitness function to optimize the size and resiliency of the SNN.  We demonstrate that this strategy leads to well-performing, small-sized networks that are more resilient to hardware faults.  
\end{abstract}

\begin{IEEEkeywords}
Neuromorphic Computing, Spiking Neural Networks, Multi-objective, Fault Tolerance, Evolutionary Optimization
\end{IEEEkeywords}

\section{Introduction}

With the advent of internet of things, smaller sensors and smarter environments are on the rise \cite{wang2017survey}. This increases the demand for low power hardware that can preform machine learning tasks \cite{samie2016iot, chatterjee2017internet}. Neuromorphic computing architectures are promising hardware architectures that fulfill these requests \cite{painkras2013spinnaker, benjamin2014neurogrid, mitchell2018danna}. These architectures are non-von Neumann chips that are inherently low power and parallel as they model a biological neural system via Spiking Neural Networks (SNNs) implementations in silicon \cite{schuman2017survey, painkras2013spinnaker, esser2015backpropagation, davies2018loihi}. However, neuromorphic hardware implementations of SNNs tend to suffer from environmental perturbations due to current or voltage variations or bit flips, among others, with the synapse weight being the most critical part susceptible to perturbations \cite{johnson2017homeostatic, sudhof2008understanding}. The perturbations that lead to a change in the synaptic parameters of the implemented spiking neural network often result in a greatly degraded performance of the network \cite{sudhof2008understanding}. 
At the same time, smaller sensors require smaller networks due to power restrictions \cite{hammoudi2018challenges}. Thus, taking neuromorphic computing to the next step and developing small size SNNs that are also resilient to perturbations of their parameters is a necessity. 

In this work we leverage an evolutionary optimization framework called EONS \cite{schuman2016evolutionary, plank2018tennlab} to train spiking neural networks. This evolutionary approach has been shown to lead to well-performing SNNs, as demonstrated on several machine learning and control tasks \cite{mitchell2018danna, reynolds2018comparison, schuman2019biomimetic}. At the core of the evolutionary (genetic) algorithm is the fitness function which evaluates every network in a population. We propose a multi-objective fitness function which incorporates a penalty for the number of neurons in a network as a way to generate smaller sized networks. We further include the performance of several variations of the network that is currently evaluated to produce networks that are resilient to some particular kind of perturbations that are possible to be encountered in the hardware. We show that this is a flexible approach for generating well performing, small sized SNNs that are more resilient to hardware faults.

\section{Background}
Reducing the size of artificial neural networks is an ongoing quest in science and technology, ever more so with the development of low-power devices for machine learning \cite{merolla2014million}. In the area of deep learning, there are many different reduction and pruning methods that aim to decrease the size of the powerful but over-parameterized deep neural networks that are deployed on traditional CPU or GPU architectures \cite{jaderberg2014speeding, chung2016simplifying, luo2017thinet, dimovska2019novel}. On the other hand, for \textit{spiking} neural networks, more efforts have been devoted to their training \cite{lee2016training, esser2015backpropagation} and not so much to their efficiency. The problems of size reduction and hardware perturbation resilience (noise sensitivity) for SNNs have been considered in limited scenarios. For example, for sparse SNNs, there are algorithms that map an SNN to a specific hardware, transforming the SNN in order to match some of the hardware constraints, such as power consumption or memory access latency \cite{ji2016neutrams, lin2019learning}. However, these methods either require sparse SNNs already or they sparsify the SNN, but there are no guarantees that the initial accuracy of the network will be preserved after the transformation.
In this work we propose a multi-objective evolutionary training method for SNNs that optimizes the network for hardware constraints such as lower power consumption and fault tolerance, while providing some confidence in the performance of the network via the training and testing accuracy. 

Multi-objective evolutionary training for neural networks has been used in several works though the objectives have been mostly centered around performance metrics or classification sensitivity or specificity \cite{yeung2015mlpnn, yuan2017multi}. Furthermore, training spiking neural networks to be fault tolerant, i.e.,  addressing certain hardware fault tolerance limitations, to the best our knowledge, is a research topic that is only beginning to emerge. For example, only a very recent work addresses the noise-handling limitations of hardware implementations of spiking neurons from a software (algorithmic) perspective \cite{olin2019stochasticity}. The work in \cite{olin2019stochasticity} adds noise to the threshold of the spiking neuron in the process of evolutionary training of the SNN, thus effectively making the trained SNN more resilient to noise fluctuations in hardware. 

In this work, we consider multi-objective fitness function in the evolutionary process to train a network that is resilient to synaptic perturbations, which are one of the most critical structures in neuromorphic hardware that are susceptible to perturbations \cite{johnson2017homeostatic, sudhof2008understanding}. The other objective in the multi-objective fitness function is to minimize the number of neurons in the SNN. We show that by penalizing for the number of neurons we can significantly reduce the number of synapses in the network as well. We demonstrate these results on two neuromorphic implementations and via several applications. We explain the genetic algorithm and the neurmorphic implementations considered in the following subsections.

\subsection{Genetic Algorithm for Training SNNs}

In this work, we use a genetic algorithm-based training approach called Evolutionary Optimization for Neuromorphic Systems (EONS) \cite{schuman2016evolutionary}. This approach evolves both the parameters of the network (e.g., weights of synapses and thresholds of neurons), as well as the structure of the network (e.g., number of neurons and synapses and how they are connected). In all of the experiments, the following parameters were used during the evolutionary process: population size of 100, crossover rate of 0.5, mutation rate of 0.9, and merge rate of 0.1. 

\subsection{DANNA2}

DANNA2 (Dynamic Adaptive Neural Network Arrays) \cite{mitchell2018danna} is a synchronous digital neuromorphic architecture with integrate-and-fire neurons and optional synaptic plasticity. Networks are represented as a directed graph in a two-dimensional space using neurons as graph nodes and synapses as edges. When a neuron's charge exceeds the configured 10-bit threshold, the neuron fires and enters a configurable refractory period in which it temporarily may not fire again. Fires from a neuron travel along synapses with an individually configurable 4-bit temporal delay and signed 9-bit weight value.

DANNA2 may be simulated on a traditional CPU or be implemented on an FPGA or ASIC. In this work, DANNA2 sparse is used which allows for a network to be converted into an optimized hardware implementation. The resulting hardware directly builds the network graph and removes any unnecessary functionality which allows for improved efficiency.

\subsection{NIDA}

Neuroscience-inspired dynamic architectures or NIDA \cite{schuman2014neuroscience} is a simple spiking neural network implementation using integrate-and-fire neurons and assumes analog synaptic weight values (specifically, floating point values between -1 and 1).  Neurons are laid out in three-dimensional space and the delays between neurons depend on the distance between neurons in the space. NIDA is implemented in simulation only. 

\section{Main Results}
In this section we present the main contribution of this article: a multi-objective fitness function that allows for the evolution of smaller-size SNNs that are also resilient to particular hardware perturbations of the synaptic weights. Another contribution of this article is the analysis of two network size-reducing strategies: during-training size optimization and post-training pruning. Namely, we provide empirical results that suggest it is better to constrain the size of a network during the training process, as opposed to pruning the network after training. We also show that penalizing for the number of neurons during the process of training actually leads to a much smaller number of synapses as well. The next subsection explores these two different size-reduction strategies. 

\subsection{Size Optimization}\FloatBarrier
There are two main approaches to reducing the size of a spiking neural network: a post-training approach and during-training approach. First, we examine the post-training approach of pruning the network by removing neurons that have low spiking frequency. The hypothesis is that those low-frequency neurons have a small contribution to the frequency of spiking of the output neurons and thus removing them will not have a significantly negative effect on the accuracy. We perform this post-training pruning analysis on three applications for the digital neuromorphic architecture DANNA2 \cite{mitchell2018danna}. The applications we consider are the following: a pole-balancing control task (abbreviated as PB), a classification task on satellite radio signal data, and an Atari-like game Asteroids. Table ~\ref{table: unconstrained stats} contains statistics from $100$ spiking neural networks that have been trained via an evolutionary algorithm which optimizes only for performance. We can see that the average spiking frequency of internal neurons is much lower than the spiking frequency of the input and output neurons. This is an indication that we could prune a large number of internal neurons from a network without a significant negative effect on performance of the network. 
 
\begin{table}[htbp]
\caption{
Statistics from $100$ spiking neural networks for digital implementation. The average spiking frequency of most of the internal neurons is an order of magnitude less than the spiking frequency of input and output neurons.}
\begin{tabular}{|l|l|l|l|}
\hline                                                                                     & PB & Radio & Asteroids \\ \hline
\begin{tabular}[c]{@{}l@{}}Avg. number \\ of internal neurons\end{tabular}                 & 9.36   & 77.8      &    75.09       \\ \hline
\begin{tabular}[c]{@{}l@{}}Avg. number\\ of synapses\end{tabular}                    & 32.61   &   139.19    & 438.18          \\ \hline
\begin{tabular}[c]{@{}l@{}}Avg. spiking freq. \\  of internal neurons\end{tabular} &  0.0015  &   0.036     &       0.00016     \\ \hline
\begin{tabular}[c]{@{}l@{}}Avg. spiking freq. \\ of input neurons\end{tabular}    & 0.033   &  0.269     &     0.00064      \\ \hline
\begin{tabular}[c]{@{}l@{}}Avg. spiking freq.\\ of output neurons\end{tabular}       &  0.022  &  0.266     &  0.0013 \\     \hline
\begin{tabular}[c]{@{}l@{}}Avg. performance \end{tabular}       &  292.2 (sec)  &  0.777 (accuracy)     &  214.9 (score) \\     \hline
\end{tabular}
\label{table: unconstrained stats}
\end{table}

Thus, we prune the internal neurons whose spiking frequency is an order of magnitude less than the spiking frequency of the output neurons. Specifically, we prune neurons whose frequency $f$ is such that $f < 0.1*f_{avg}(\text{outputs})$, where $f_{avg}(\text{outputs})$ denotes the average spiking frequency of the output neurons. When performing such pruning, we get networks that have most of the internal neurons of the original network pruned, as shown in Table~\ref{table: pruning stats}. Though the number of internal neurons and the number of synapses is very significantly decreased, the loss of performance is comparatively small, as we can see from Table~\ref{table: pruning stats}.

\begin{table}[htbp]
\caption{
Statistics from $100$ spiking neural networks for digital implementation that were pruned based on low frequency spiking internal neurons. }
\begin{tabular}{|l|l|l|l|}
\hline                                                                                     & PB & Radio & Asteroids \\ \hline
\begin{tabular}[c]{@{}l@{}}Avg. number \\ of internal neurons\end{tabular}                 & 2.43   & 4.97     &    35.24       \\ \hline
\begin{tabular}[c]{@{}l@{}}Avg. number\\ of synapses\end{tabular}                    & 21.53   &   62.52    & 393.4          \\ \hline
\begin{tabular}[c]{@{}l@{}}Avg. Performance\end{tabular}                    & 288.7 (sec)   &   0.778 (accuracy)    & 208.3 (score)          \\ \hline
\end{tabular}
\label{table: pruning stats}
\end{table}

These results lead to the question of whether it is worth implementing a penalty for the size of the network during the training process, or whether post-training pruning performs better, i.e. leads to a smaller size and higher performance of the network. To answer this, we generate $100$ networks from the same applications, but in this case, during the evolutionary optimization training of these networks, we penalize the fitness function for the number of neurons. In particular, the fitness function $F(N)$ for a network $N$ that we consider is the following: 

 \begin{align*}
        F(N) &= \text{performance}(N)* \\
        &*(1-\frac{\text{num. of hidden neurons}(N)}{\text{total number of neurons}(N)}*\delta)\,,
\end{align*}
where $\delta=0.001$ has been experimentally chosen. 
Evolving networks in this manner we get the best results for both size and performance of the networks, as shown in Table~\ref{table: constrained fitness1 stats}. We also note that the number of synapses in the resulting networks is also very low. 

\begin{table}[htbp]
\caption{
Statistics from $100$ spiking neural networks for digital implementation that were evolved with a multi-objective fitness that optimizes for performance and for the number of neurons in the network. 
}
\begin{tabular}{|l|l|l|l|}
\hline                                                                                     & PB & Radio & Asteroids \\ \hline
\begin{tabular}[c]{@{}l@{}}Avg. number \\ of internal neurons\end{tabular}                 & 5.08   & 0.56     &   1.3       \\ \hline
\begin{tabular}[c]{@{}l@{}}Avg. number\\ of synapses\end{tabular}                    &23.41  &   122.9   & 147.17        \\ \hline
\begin{tabular}[c]{@{}l@{}}Avg. performance \end{tabular}
    & 297.5 (sec) &   0.788 (accuracy)   & 235.22 (score)        \\ \hline
\end{tabular}
\label{table: constrained fitness1 stats}

\end{table}
Furthermore, as we can see from Table~\ref{table: constrained fitness1 stats}, using a multi-objective fitness function to generate smaller sized networks even improves the average performance of the networks. Hence, we keep the size-penalty in the fitness function in our further experiments and we choose to optimize networks for size during training instead of the post-training pruning. We note however that we observed longer training time with the multi-objective fitness function, thus if computing resources for training are limited, the post-training pruning approach is a good alternative for decreasing the size of a network. 

\subsection{Size and Resilience Optimization}\FloatBarrier

\begin{figure}
	\centering
	\bgroup
	\def\arraystretch{0.6}
	\begin{tabular}{c}
	\includegraphics[width=0.9\columnwidth, keepaspectratio]{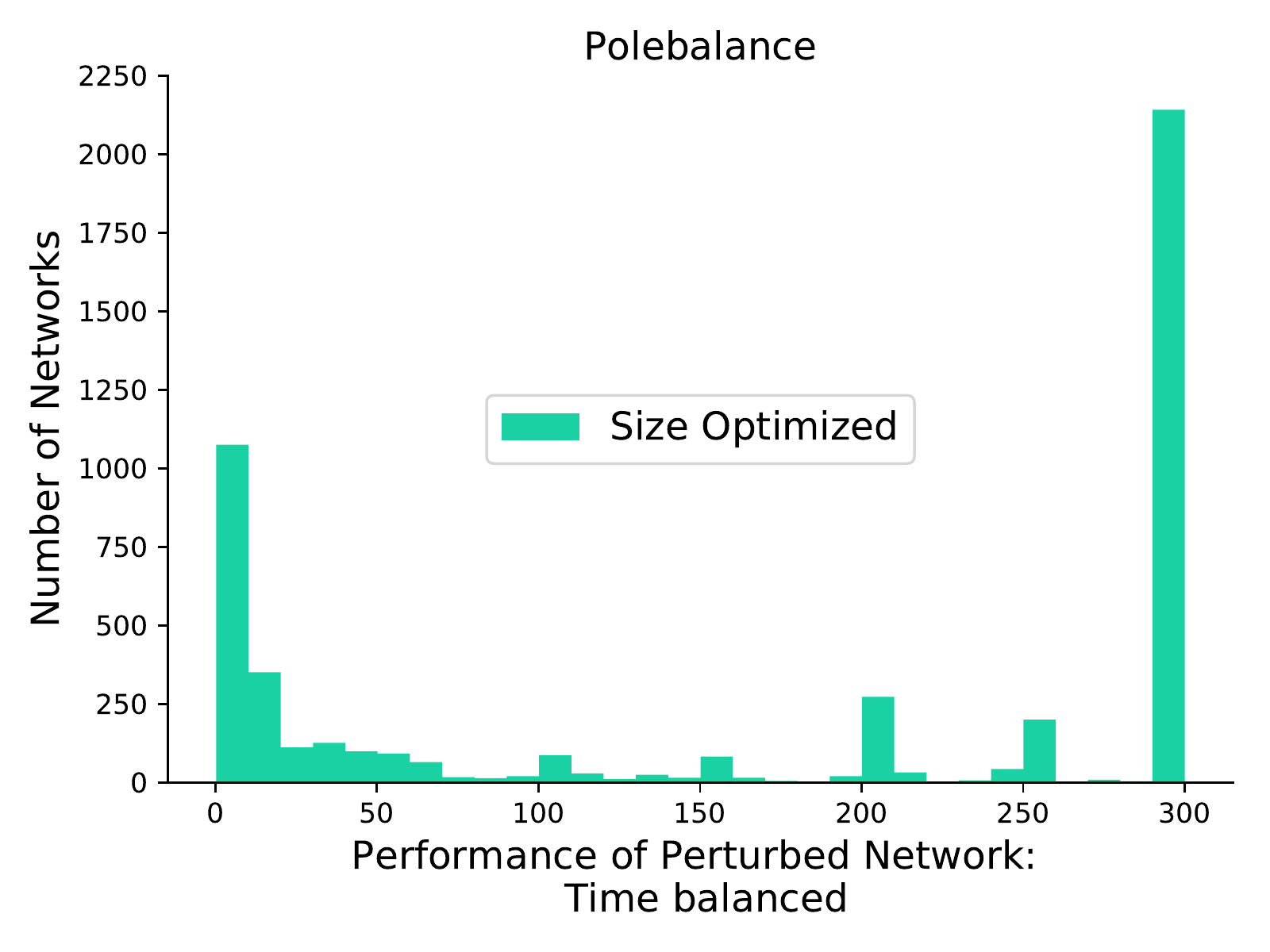} \\
	(a)\\
	\includegraphics[width=0.9\columnwidth, keepaspectratio]{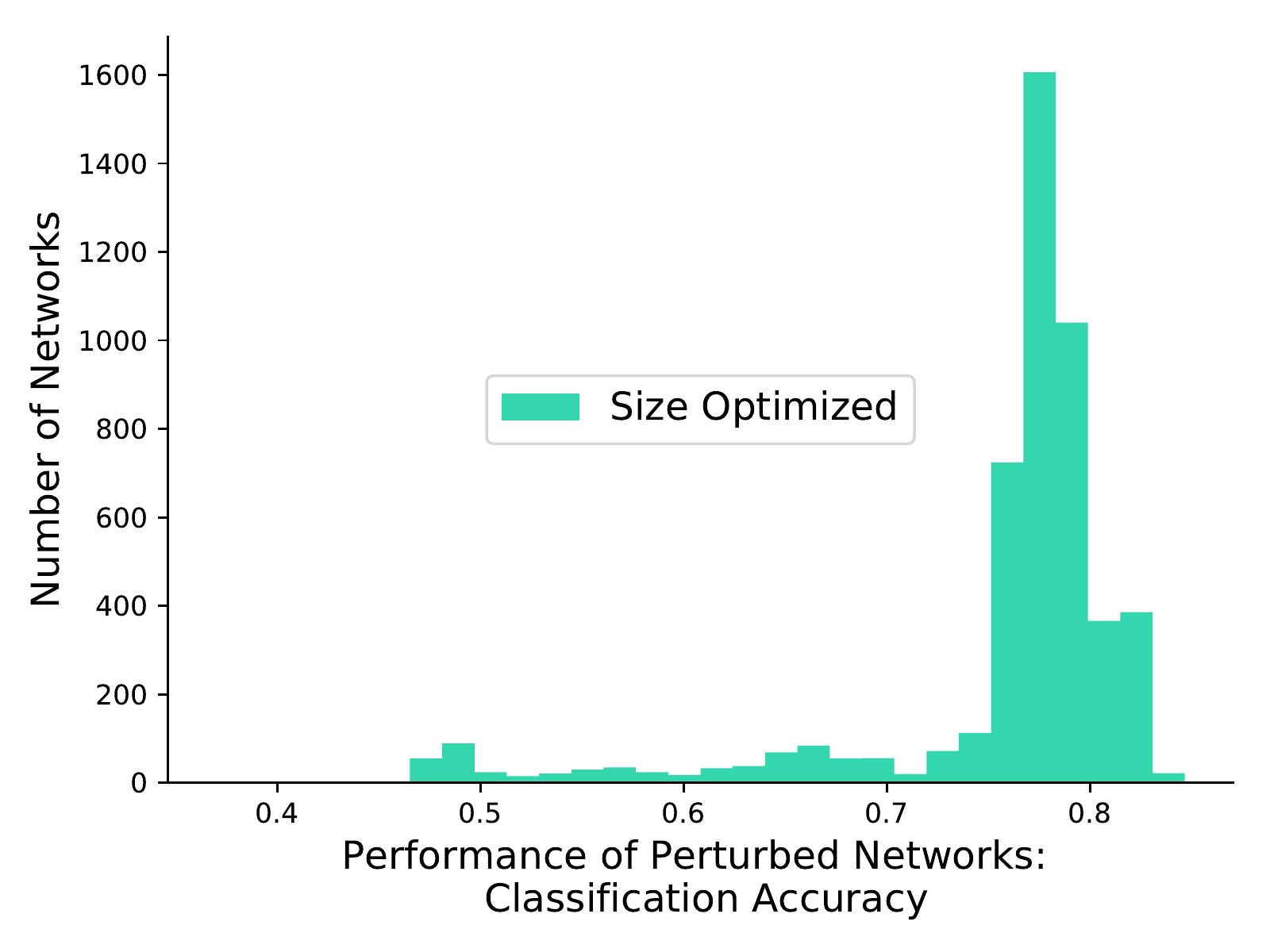} \\
	(b) \\
    \end{tabular}
	\egroup
	\caption{Perturbing $10$ of the size-optimized networks by flipping their $8^\text{th}$ bit. (a) Pole balance application: many of the synapses are very sensitive to this perturbation as more than a third of the resulting generated (perturbed) networks have their balancing time degrading to zero. (b) Radio application: though the SNNs for this application are less sensitive to perturbations, there is still a significant percentage of synapses whose perturbation leads to loss of accuracy. \label{fig:inout sensitivity}}
\end{figure}

Though optimizing for size and performance during training gave better results than pruning the network post-training, because the number of internal neurons is very low, the resulting size-optimized networks have many input to output synapses. We performed a simple perturbation experiment to test the resiliency to synaptic weight perturbations of these size-optimized networks. Namely, we randomly chose $10$ networks out of those $100$ size-optimized networks from each of the pole balance and the radio application networks. We perturbed those networks by randomly sampling $1, 2, 3, 4 $ and $5$ of its synapses for $100$ times consecutively and flipping one bit in the weight of the sampled synapse. Thus, we get $5000$ perturbed networks in total. We chose the $8^{th}$ bit from the sampled synapses' weights to be flipped. Having networks that are resilient to such a bit flip change or similar changes is especially important for sensors that are in an environment with high radioactivity, where, for example, bit flip resilience is crucial to the sensors reliability \cite{elliott2016exploiting}. But the size and performance optimized networks are observed to be sensitive to such perturbations. Namely, in the experiment many of the resulted perturbed networks have their performance significantly affected, as shown in Figure~\ref{fig:inout sensitivity}. Namely, $35.3\%$ of the perturbed networks for the pole balance task have balancing time of less than $50$ sec, while the optimal balancing time for this app is $300$ sec. For the radio application, the evolutionary algorithm  generates networks that are much more resilient to bit flips when compared to the SNNs for the pole balance application, but here too we have that almost $7\%$ of the perturbed networks have accuracy less than $60\%$, while the un-perturbed size-optimized networks all had accuracy greater than $60\%$. Ultimately, we'd like the percentage of perturbed networks that lose their good performance to be as small as possible. 

Given these observations we propose the following multi-objective fitness function $F(N)$ for spiking neural network $N$, that penalizes both for the number of neurons and for the sensitivity to perturbations:

 \begin{align*}
       F(N) &= w_{1}*\text{performance}(N)*\\
       &*(1-\frac{\text{num. of hidden neurons}}{\text{total number of neurons}}*\delta) + \\
        & + w_{2}*(\displaystyle \frac{\sum_{i=1}^{n}{\text{performance of variation}(N_{i})}}{n})
\end{align*}\label{eq: MO fitness}
 where $\delta, n, w_{1}, w_{2}$ are parameters that are to be chosen experimentally with the following constraints: $w_{1}, w_{2} \in [0,1]$ and $w_{1} + w_{2} = 1$; $\delta \in (0,1)$; $n \geq 1$. 
By $N_{i}$ we denote a variation of the network whose fitness is currently evaluated. The variation will depend on the type of perturbation that we are expecting to see in hardware. For example, if we are considering a digital implementation of a SNN, then a typical hardware fault is a bit flip. Then, one type of variation of the network that we can consider is a network that has one (or more) of its synapse weights changed to a weight that has one (or more) of its bits flipped. 

\subsection{Multi-objective Evolutionary Optimization Applications}

We apply the multi-objective function that optimizes for size and resilience to applications from two types of neuromorphic implementations: one where the weights are integer values and one which admits floating point weights. Details about the digital implementation with integer weights (DANNA2) can be found in \cite{mitchell2018danna} and details about the floating-point weights implementation (NIDA) can be found in \cite{schuman2014neuroscience}. The main distinction of these two implementations for the purposes of this work, is that in the digital implementation the synaptic weights are integers in the range $[-1024, 1024]$ and in the NIDA implementation synapses admit floating point weights between $-1$ and $1$.  
In all of our experiments, the parameters for the multi-objective fitness function stated in Equation~\ref{eq: MO fitness} were experimentally chosen as $\delta=0.001$ $n=5$, and $w_{1} = w_{2} = 0.5$. For the Pole balance task, all the networks were trained until they achieved training balancing time of $300.02$ seconds; for the radio task, all the networks were trained until they achieved accuracy of $77\%$ or greater; for the asteroids app, all the networks were trained for $50$ epochs. While for the resiliency-and-size optimized networks we achieve better fault tolerance, the training time was naturally longer than if we were optimizing networks for size and performance only. The experiments are further detailed below.

\subsection*{\bf DANNA2 Pole Balance}
In a digital implementation, the type of hardware fault that can be experienced is a bit flip. As flipping the $8^{th}$ bit of digitally implemented SNNs led to significant loss of performance in the performance-and-size only optimized networks, we considered evolving networks that are resilient to the flip of the $8^{th}$ bit. To this end, we considered variations in which every synapse has the $8^{th}$ bit flipped with probability $0.1$. In summary, the SNN synaptic weight operations considered were the following: 
\begin{itemize}
    \item Perturbation: a sampled synapse has $8^{th}$ bit flipped. 
    \item Variations: $5$ variations were considered, and in each network variation, each synapse has its  $8^{th}$ bit flipped with probability $0.1$.
\end{itemize}

\begin{figure}
	\centering
	\bgroup
	\def\arraystretch{0.6}
	\begin{tabular}{c}
	\includegraphics[width=0.9\columnwidth, keepaspectratio]{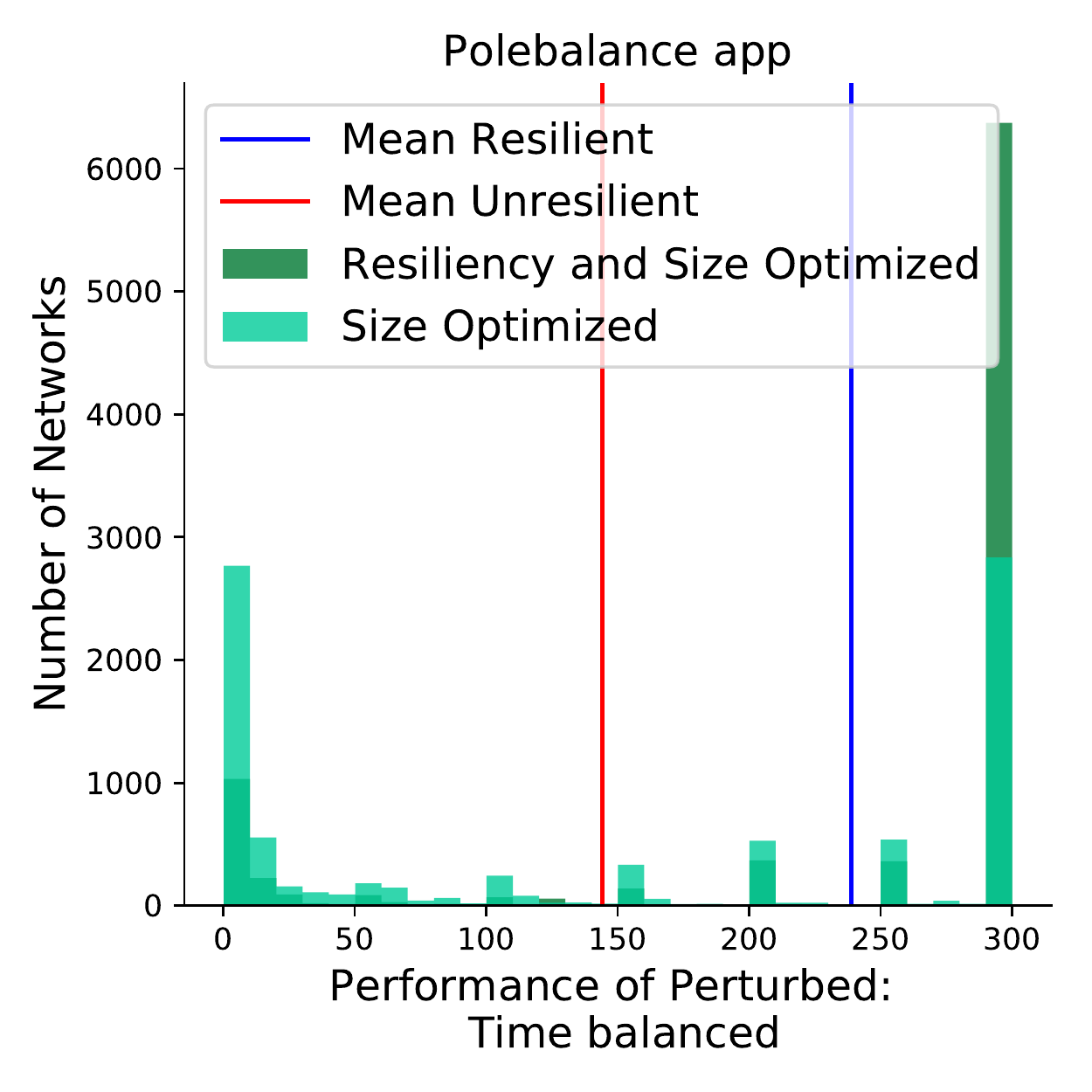} \\
	(a)\\
	\includegraphics[width=0.9\columnwidth, keepaspectratio]{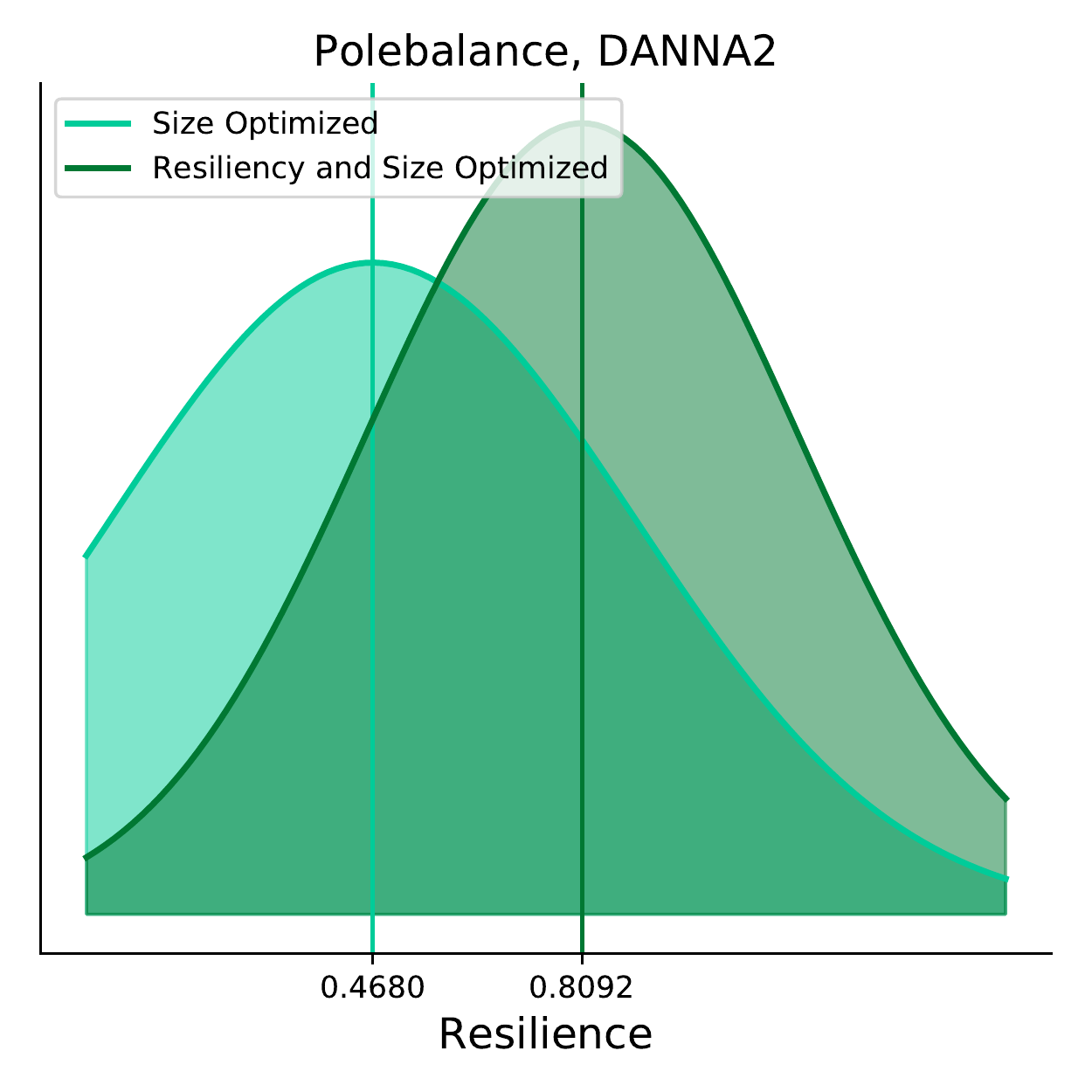} \\
	(b) \\
    \end{tabular}
	\egroup
	\caption{Perturbing $20$ size-optimized and $20$ size-and-resilience optimized networks by flipping their $8^\text{th}$ bit: we sample $1$ through $5$ synapses for $100$ times for each of those networks and flip the  $8^\text{th}$ bit of the sampled synapses. (a) Histogram of the experiment. (b) Fitted Gaussians from resiliency scores.  \label{fig:danna2_bit_flip}}
\end{figure}

As we can see from Figure~\ref{fig:danna2_bit_flip}, evolving the networks with the resilience to perturbations taken into account in the fitness function leads to much more resilient networks (fewer sensitive synapses). In Figure~\ref{fig:danna2_bit_flip}(b), a Gaussian is fitted for both the resilience metric of the resiliency optimized and the resilience unoptimized networks. The resilience metric for each network $N$ that we calculated is:
\begin{equation}
\frac{\text{optimal performance} - \text{network performance}}{\text{optimal performance}}
\end{equation}

 where for the pole balance task, optimal performance is $300.02$ seconds. As seen in Figure~\ref{fig:danna2_bit_flip}(b), the resilience-optimized networks have a twice higher resiliency mean than the size-and-performance only optimized SNNs. 


\subsection*{\bf NIDA Polebalance}
\begin{figure}
	\centering
	\bgroup
	\def\arraystretch{0.6}
	\begin{tabular}{c}
	\includegraphics[width=0.9\columnwidth, keepaspectratio]{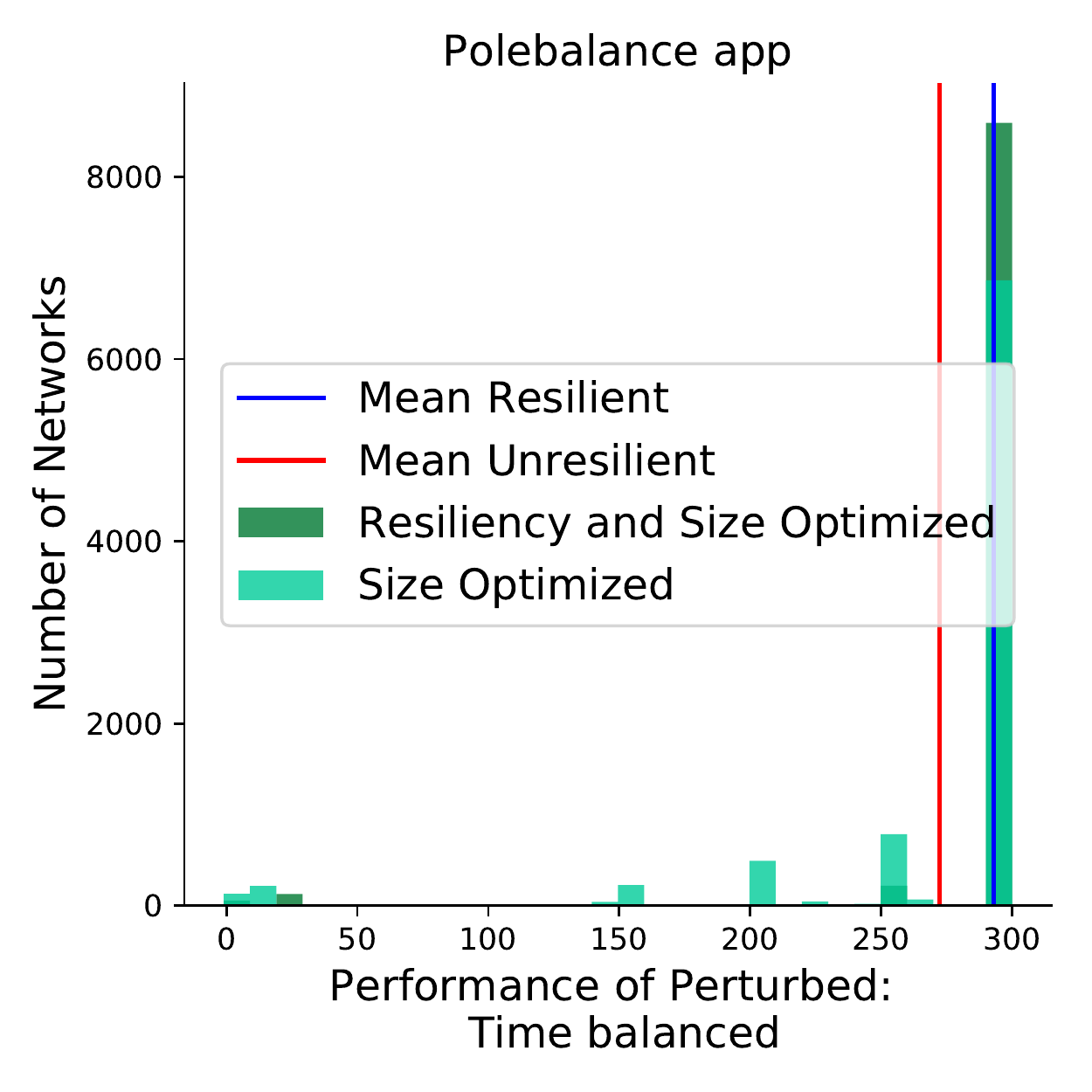} \\
    \end{tabular}
	\egroup
	\caption{Perturbing $20$ of the size-optimized and $20$ size-and-resilience optimized networks by diminishing them for $\epsilon \in (0.005, 0.05)$: we sample $1$ through $5$ synapses for $100$ times for each of those networks and diminish the sampled synapses. The size-and-resilience optimized networks have higher mean (almost all of them balance at around $300$ sec., which is not the case for the size-only optimized networks).  \label{fig:pb_nida}}
\end{figure}

In the NIDA implementation, the type of hardware fault that can be experienced is a ``diminishing weight''. Namely, due to current and voltage drops, the weight value can diminish closer to zero. Thus, the synaptic weight operations considered were the following: 
\begin{itemize}
      \item Perturbation: a sampled synapse is diminished by $\epsilon \in (0.005, 0.05)$. 
     \item For the multi-objective fitness function, $5$ variations are considered for resilience. In each variation, every synapse is sampled with probability $0.5$ and is diminished by $0.05$.
\end{itemize}
The results, showing that the resilience optimized networks are indeed resilient to these perturbations, while the size-optimized networks are not, are shown in Figure~\ref{fig:pb_nida}. 

\subsection*{\bf NIDA Radio}
The performance-and-size-optimized networks for the Radio signal classification application were more resilient to diminishing weight changes, as most of the networks perturbed in such a way still led to classification greater than $75\%$. However, the networks are less resilient to decrements by  $\epsilon \in (0.008, 0.1)$. Thus, the synaptic weight operations considered were the following: 
\begin{itemize}
    \item A sampled synapse is decreased by $\epsilon \in (0.008, 0.1)$. 
    \item For the multiobjective fitness function, $5$ variations for resilience were considered. In each variation, every synapse is decreased by $\epsilon \in (0.008, 0.1)$, where $\epsilon$ is sampled uniformly from the interval $(0.008, 0.1)$.
\end{itemize}

\begin{figure}
	\centering
	\bgroup
	\def\arraystretch{0.6}
	\begin{tabular}{c}
	\includegraphics[width=0.9\columnwidth, keepaspectratio]{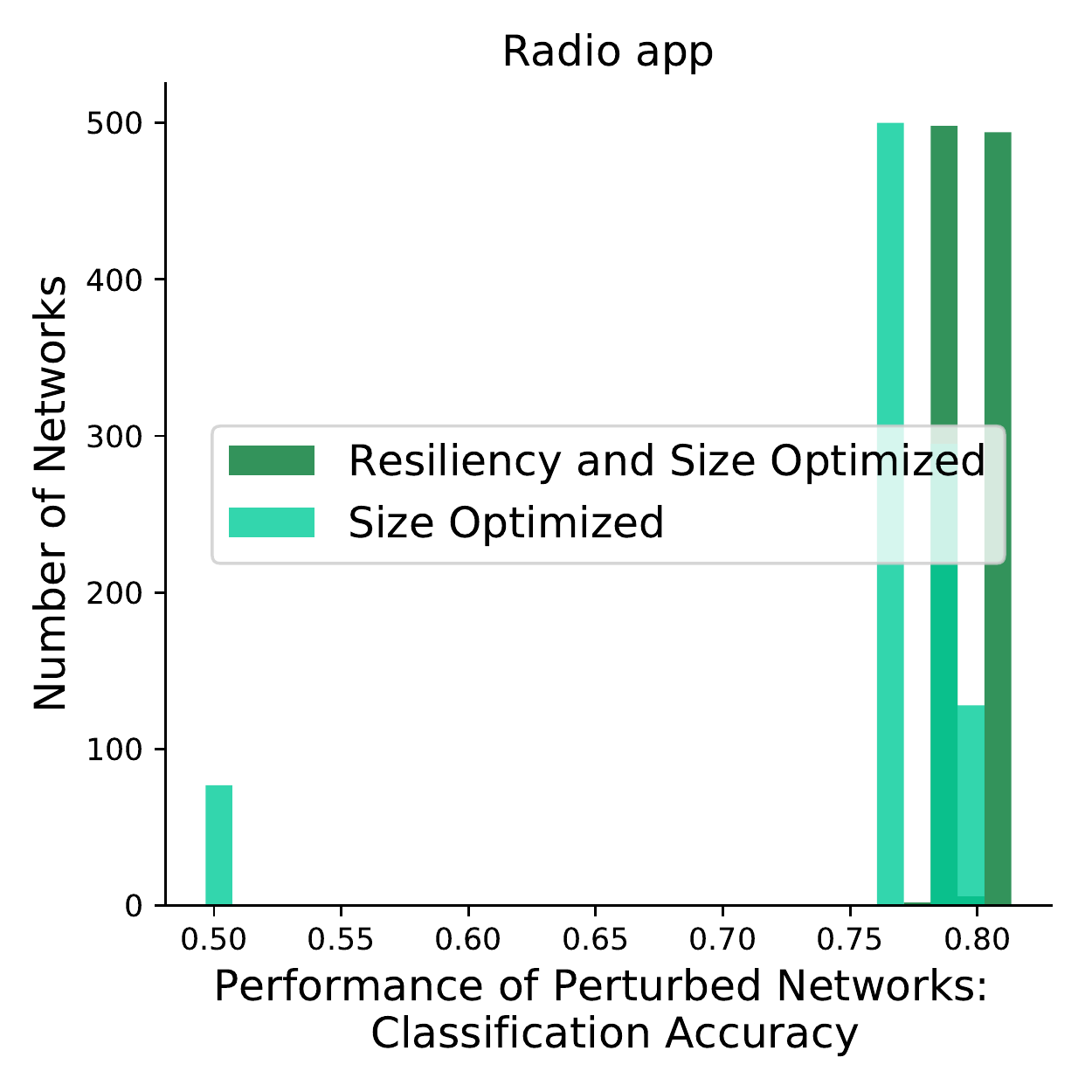} \\
    \end{tabular}
	\egroup
	\caption{Perturbing $5$ of the size-optimized and size-and-resilience optimized networks by decreasing them for $\epsilon \in (0.008, 0.1)$: we sample $1$ through $5$ synapses for $100$ times for each of those networks and diminish the sampled synapses. The size-and-resilience optimized networks all retain their accuracy when perturbed (they perform at around $75-80\%$ classification accuracy), while a portion of the perturbed size and performance-only optimized networks has accuracy around $50\%$. \label{fig:radio_nida}}
\end{figure}

The results are show in Figure~\ref{fig:radio_nida}. Again, the negative effect on performance in the resilience optimized perturbed networks is not present. 

\subsection*{\bf NIDA Asteroids}
For the asteroid application, we perform a perturbation that diminishes the synaptic weights by $\epsilon \in (0.001, 0.1)$. In this application we have three metrics: the time the player stayed alive, its score, and the total shooting points the player has acquired. We chose the score as a comparison metric between the networks that were optimized for size only and the networks optimized for size and resilience. We have trained $10$ networks for $50$ epochs in both scenarios. The experiments involved the following synaptic operations:

\begin{itemize}
    \item Perturbation: a sampled synapse is diminished by $\epsilon \in (0.001, 0.1)$. 
    \item Variations: in the multiobjective fitness, $5$ variations are considered. For every synapse in each variation, we diminish its weight by $\epsilon = 0.1$ with probability $0.5$. 
\end{itemize}

\begin{figure}
	\centering
	\bgroup
	\def\arraystretch{0.6}
	\begin{tabular}{c}
	\includegraphics[width=0.9\columnwidth,  keepaspectratio]{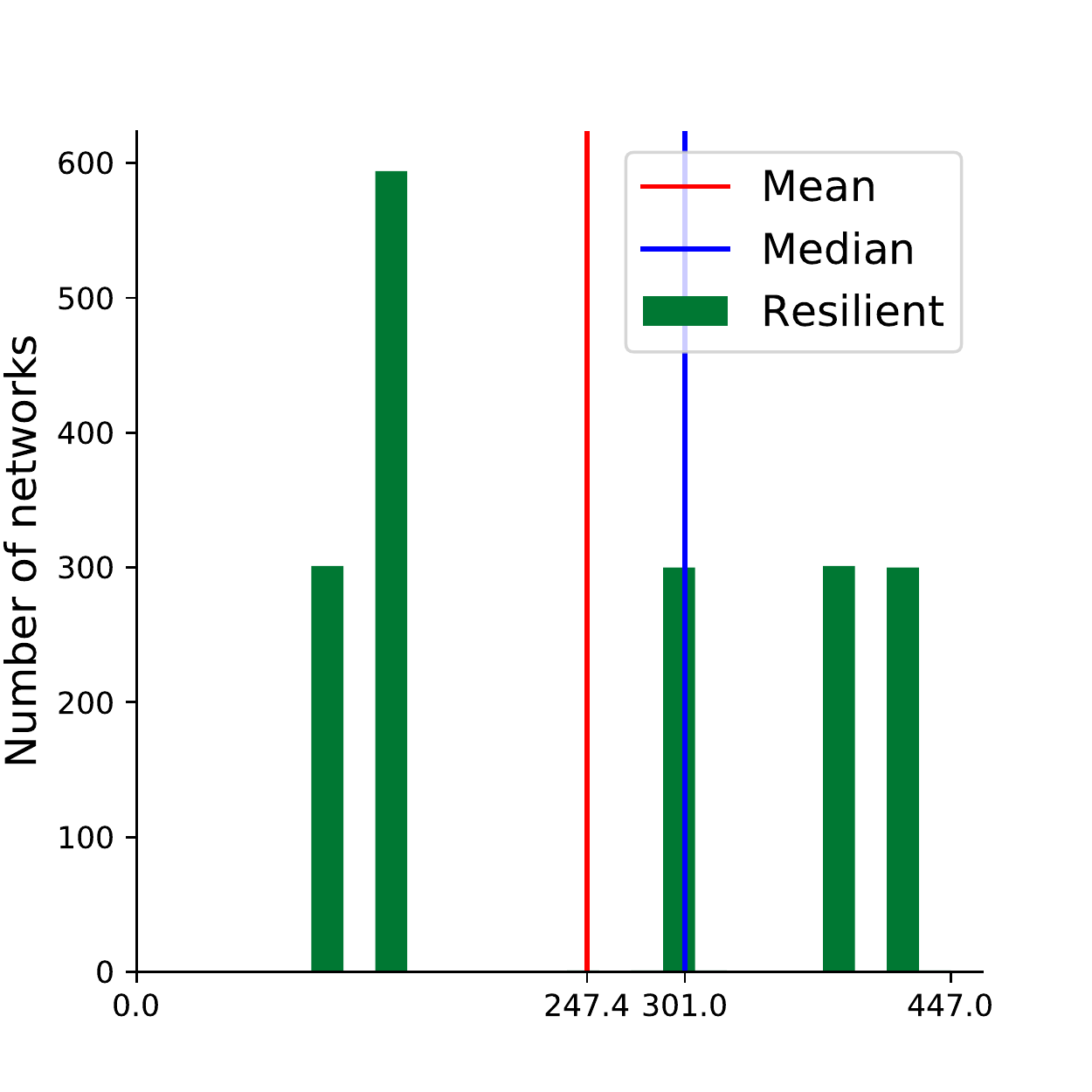} \\
	(a)\\
	\includegraphics[width=0.9\columnwidth, keepaspectratio]{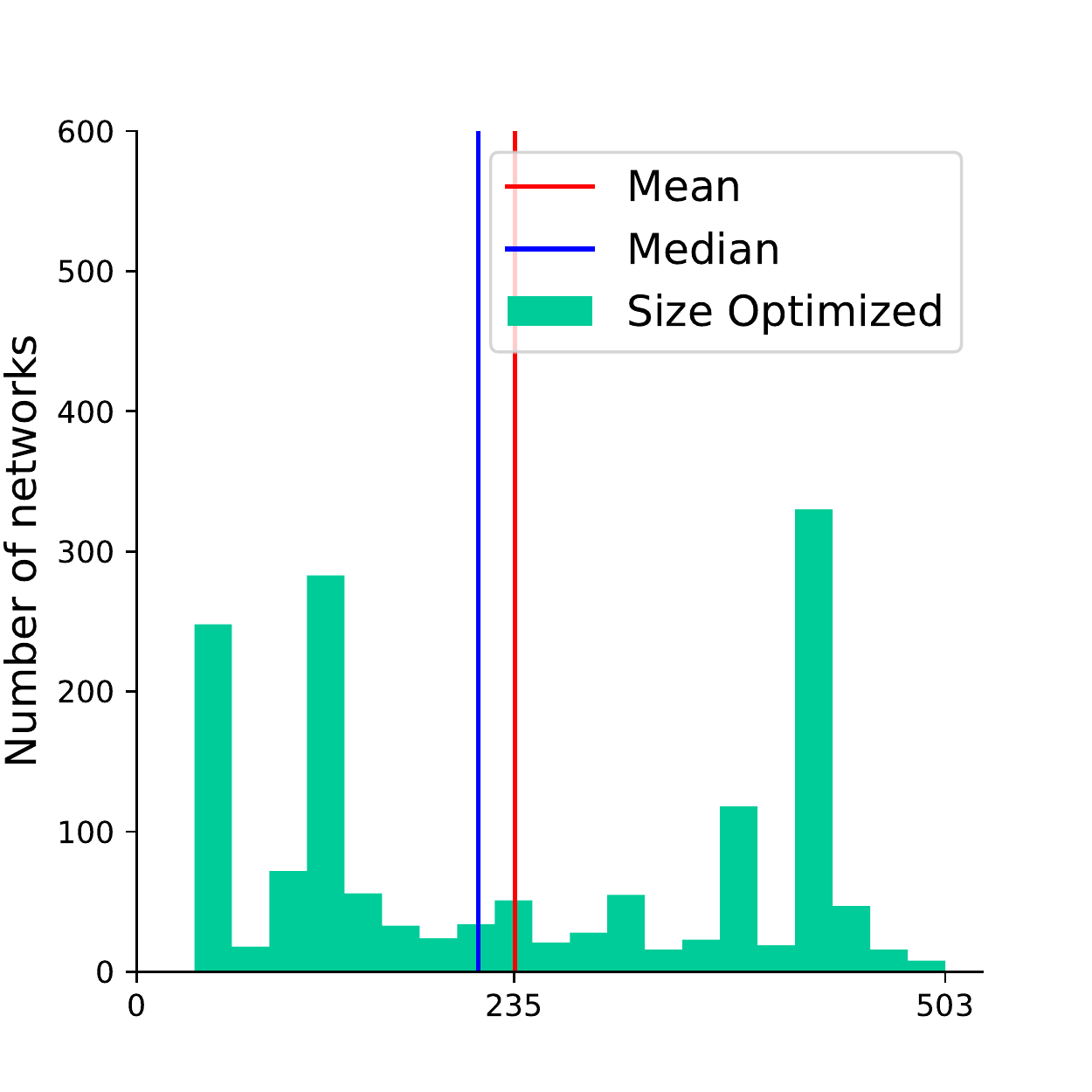}\\
	(b)
    \end{tabular}
	\egroup
	\caption{Perturbing $10$ of the size-optimized and size-and-resilience optimized networks by diminishing them by $\epsilon \in (0.001, 0.1)$: we sample $1$ through $5$ synapses for $100$ times for each of those networks and diminish the sampled synapses. The size-and-resilience optimized networks show better performance when perturbed: they have higher mean and median for the score of perturbed networks. (a) Performance of perturbed networks when networks were evolved with performance, size and perturbation resilience optimization. (b) Performance of perturbed networks when networks were trained with performance and size optimization. \label{fig:ast_nida}}
\end{figure}

The results are show in Figure~\ref{fig:ast_nida}. Again, the resiliency optimized networks show higher mean and median in performance when perturbed than the size and performance optimized SNNs.   

\section{Conclusion and Discussion}

In this work we propose a multi-objective fitness optimization function for training Spiking Neural Networks via an evolutionary algorithm. This function optimizes for performance, size and perturbation resilience of the network. The factor that accounts for perturbation resilience involves several variations of the network whose fitness is evaluated and takes into account their performance. The network variations that we consider depend on the type of perturbation we can encounter due to hardware faults. 
Another contribution of this article is an empirical comparison of two size-reduction methods, where we show that in-training size constraints lead to better results in terms of network size and accuracy as opposed to pruning networks post-training. In summary, this work shows that algorithmic, software-side solutions to producing spiking neural networks which are resilient to hardware faults and satisfy hardware constraints are possible. As future work, it would be interesting to consider finding good initialization techniques for the seed networks in the evolutionary algorithm as well as performing a hyperparameter analysis of the parameters in the multi-objective fitness function. This would allow for a faster convergence to well-performing and resilient networks, trained by utilizing the multi-objective fitness function. 

\section*{Acknowledgment}
This material is based upon work supported by the U.S. Department of Energy, Office of Science, Office of Advanced Scientific Computing Research, Robinson Pino, program manager, under contract number DE-AC05-00OR22725, and in part by the Laboratory
Directed Research and Development Program of Oak
Ridge National Laboratory.

\bibliographystyle{IEEEtran}
\bibliography{sample-base}

\begin{thebibliography}{10}
\providecommand{\url}[1]{#1}
\csname url@samestyle\endcsname
\providecommand{\newblock}{\relax}
\providecommand{\bibinfo}[2]{#2}
\providecommand{\BIBentrySTDinterwordspacing}{\spaceskip=0pt\relax}
\providecommand{\BIBentryALTinterwordstretchfactor}{4}
\providecommand{\BIBentryALTinterwordspacing}{\spaceskip=\fontdimen2\font plus
\BIBentryALTinterwordstretchfactor\fontdimen3\font minus
  \fontdimen4\font\relax}
\providecommand{\BIBforeignlanguage}[2]{{%
\expandafter\ifx\csname l@#1\endcsname\relax
\typeout{** WARNING: IEEEtran.bst: No hyphenation pattern has been}%
\typeout{** loaded for the language `#1'. Using the pattern for}%
\typeout{** the default language instead.}%
\else
\language=\csname l@#1\endcsname
\fi
#2}}
\providecommand{\BIBdecl}{\relax}
\BIBdecl

\bibitem{wang2017survey}
X.~Wang, H.~Qiu, and F.~Xie, ``A survey on the industrial readiness for
  internet of things,'' in \emph{2017 IEEE 8th Annual Ubiquitous Computing,
  Electronics and Mobile Communication Conference (UEMCON)}.\hskip 1em plus
  0.5em minus 0.4em\relax IEEE, 2017, pp. 591--596.

\bibitem{samie2016iot}
F.~Samie, L.~Bauer, and J.~Henkel, ``Iot technologies for embedded computing: A
  survey,'' in \emph{Proceedings of the Eleventh IEEE/ACM/IFIP International
  Conference on Hardware/Software Codesign and System Synthesis}.\hskip 1em
  plus 0.5em minus 0.4em\relax ACM, 2016, p.~8.

\bibitem{chatterjee2017internet}
S.~Chatterjee, S.~Chatterjee, S.~Choudhury, S.~Basak, S.~Dey, S.~Sain, K.~S.
  Ghosal, N.~Dalmia, and S.~Sircar, ``Internet of things and body area
  network-an integrated future,'' in \emph{2017 IEEE 8th Annual Ubiquitous
  Computing, Electronics and Mobile Communication Conference (UEMCON)}.\hskip
  1em plus 0.5em minus 0.4em\relax IEEE, 2017, pp. 396--400.

\bibitem{painkras2013spinnaker}
E.~Painkras, L.~A. Plana, J.~Garside, S.~Temple, F.~Galluppi, C.~Patterson,
  D.~R. Lester, A.~D. Brown, and S.~B. Furber, ``Spinnaker: A 1-w 18-core
  system-on-chip for massively-parallel neural network simulation,'' \emph{IEEE
  Journal of Solid-State Circuits}, vol.~48, no.~8, pp. 1943--1953, 2013.

\bibitem{benjamin2014neurogrid}
B.~V. Benjamin, P.~Gao, E.~McQuinn, S.~Choudhary, A.~R. Chandrasekaran, J.-M.
  Bussat, R.~Alvarez-Icaza, J.~V. Arthur, P.~A. Merolla, and K.~Boahen,
  ``Neurogrid: A mixed-analog-digital multichip system for large-scale neural
  simulations,'' \emph{Proceedings of the IEEE}, vol. 102, no.~5, pp. 699--716,
  2014.

\bibitem{mitchell2018danna}
J.~P. Mitchell, M.~E. Dean, G.~R. Bruer, J.~S. Plank, and G.~S. Rose, ``Danna
  2: Dynamic adaptive neural network arrays,'' in \emph{Proceedings of the
  International Conference on Neuromorphic Systems}.\hskip 1em plus 0.5em minus
  0.4em\relax ACM, 2018, p.~10.

\bibitem{schuman2017survey}
C.~D. Schuman, T.~E. Potok, R.~M. Patton, J.~D. Birdwell, M.~E. Dean, G.~S.
  Rose, and J.~S. Plank, ``A survey of neuromorphic computing and neural
  networks in hardware,'' \emph{arXiv preprint arXiv:1705.06963}, 2017.

\bibitem{esser2015backpropagation}
S.~K. Esser, R.~Appuswamy, P.~Merolla, J.~V. Arthur, and D.~S. Modha,
  ``Backpropagation for energy-efficient neuromorphic computing,'' in
  \emph{Advances in Neural Information Processing Systems}, 2015, pp.
  1117--1125.

\bibitem{davies2018loihi}
M.~Davies, N.~Srinivasa, T.-H. Lin, G.~Chinya, Y.~Cao, S.~H. Choday, G.~Dimou,
  P.~Joshi, N.~Imam, S.~Jain \emph{et~al.}, ``Loihi: A neuromorphic manycore
  processor with on-chip learning,'' \emph{IEEE Micro}, vol.~38, no.~1, pp.
  82--99, 2018.

\bibitem{johnson2017homeostatic}
A.~P. Johnson, J.~Liu, A.~G. Millard, S.~Karim, A.~M. Tyrrell, J.~Harkin,
  J.~Timmis, L.~J. McDaid, and D.~M. Halliday, ``Homeostatic fault tolerance in
  spiking neural networks: A dynamic hardware perspective,'' \emph{IEEE
  Transactions on Circuits and Systems I: Regular Papers}, vol.~65, no.~2, pp.
  687--699, 2017.

\bibitem{sudhof2008understanding}
T.~C. S{\"u}dhof and R.~C. Malenka, ``Understanding synapses: past, present,
  and future,'' \emph{Neuron}, vol.~60, no.~3, pp. 469--476, 2008.

\bibitem{hammoudi2018challenges}
S.~Hammoudi, Z.~Aliouat, and S.~Harous, ``Challenges and research directions
  for internet of things,'' \emph{Telecommunication Systems}, vol.~67, no.~2,
  pp. 367--385, 2018.

\bibitem{schuman2016evolutionary}
C.~D. Schuman, J.~S. Plank, A.~Disney, and J.~Reynolds, ``An evolutionary
  optimization framework for neural networks and neuromorphic architectures,''
  in \emph{2016 International Joint Conference on Neural Networks
  (IJCNN)}.\hskip 1em plus 0.5em minus 0.4em\relax IEEE, 2016, pp. 145--154.

\bibitem{plank2018tennlab}
J.~S. Plank, C.~D. Schuman, G.~Bruer, M.~E. Dean, and G.~S. Rose, ``The tennlab
  exploratory neuromorphic computing framework,'' \emph{IEEE Letters of the
  Computer Society}, vol.~1, no.~2, pp. 17--20, 2018.

\bibitem{reynolds2018comparison}
J.~J. Reynolds, J.~S. Plank, C.~D. Schuman, G.~R. Bruer, A.~W. Disney, M.~E.
  Dean, and G.~S. Rose, ``A comparison of neuromorphic classification tasks,''
  in \emph{Proceedings of the International Conference on Neuromorphic
  Systems}.\hskip 1em plus 0.5em minus 0.4em\relax ACM, 2018, p.~12.

\bibitem{schuman2019biomimetic}
C.~D. Schuman, J.~S. Najem, R.~Weiss, N.~Skuda, A.~Belianinov, P.~Collier,
  S.~Sarles, G.~Rose \emph{et~al.}, ``Biomimetic, soft-material synapse for
  neuromorphic computing: from device to network,'' Oak Ridge National
  Lab.(ORNL), Oak Ridge, TN (United States), Tech. Rep., 2019.

\bibitem{merolla2014million}
P.~A. Merolla, J.~V. Arthur, R.~Alvarez-Icaza, A.~S. Cassidy, J.~Sawada,
  F.~Akopyan, B.~L. Jackson, N.~Imam, C.~Guo, Y.~Nakamura \emph{et~al.}, ``A
  million spiking-neuron integrated circuit with a scalable communication
  network and interface,'' \emph{Science}, vol. 345, no. 6197, pp. 668--673,
  2014.

\bibitem{jaderberg2014speeding}
M.~Jaderberg, A.~Vedaldi, and A.~Zisserman, ``Speeding up convolutional neural
  networks with low rank expansions,'' \emph{arXiv preprint arXiv:1405.3866},
  2014.

\bibitem{chung2016simplifying}
J.~Chung and T.~Shin, ``Simplifying deep neural networks for neuromorphic
  architectures,'' in \emph{2016 53nd ACM/EDAC/IEEE Design Automation
  Conference (DAC)}.\hskip 1em plus 0.5em minus 0.4em\relax IEEE, 2016, pp.
  1--6.

\bibitem{luo2017thinet}
J.-H. Luo, J.~Wu, and W.~Lin, ``Thinet: A filter level pruning method for deep
  neural network compression,'' in \emph{Proceedings of the IEEE international
  conference on computer vision}, 2017, pp. 5058--5066.

\bibitem{dimovska2019novel}
M.~Dimovska and T.~Johnston, ``A novel pruning method for convolutional neural
  networks based off identifying critical filters,'' in \emph{Proceedings of
  the Practice and Experience in Advanced Research Computing on Rise of the
  Machines (learning)}.\hskip 1em plus 0.5em minus 0.4em\relax ACM, 2019,
  p.~63.

\bibitem{lee2016training}
J.~H. Lee, T.~Delbruck, and M.~Pfeiffer, ``Training deep spiking neural
  networks using backpropagation,'' \emph{Frontiers in neuroscience}, vol.~10,
  p. 508, 2016.

\bibitem{ji2016neutrams}
Y.~Ji, Y.~Zhang, S.~Li, P.~Chi, C.~Jiang, P.~Qu, Y.~Xie, and W.~Chen,
  ``Neutrams: Neural network transformation and co-design under neuromorphic
  hardware constraints,'' in \emph{The 49th Annual IEEE/ACM International
  Symposium on Microarchitecture}.\hskip 1em plus 0.5em minus 0.4em\relax IEEE
  Press, 2016, p.~21.

\bibitem{lin2019learning}
J.~Lin, Z.~Zhu, Y.~Wang, and Y.~Xie, ``Learning the sparsity for reram: mapping
  and pruning sparse neural network for reram based accelerator,'' in
  \emph{Proceedings of the 24th Asia and South Pacific Design Automation
  Conference}.\hskip 1em plus 0.5em minus 0.4em\relax ACM, 2019, pp. 639--644.

\bibitem{yeung2015mlpnn}
D.~S. Yeung, J.-C. Li, W.~W. Ng, and P.~P. Chan, ``Mlpnn training via a
  multiobjective optimization of training error and stochastic sensitivity,''
  \emph{IEEE transactions on neural networks and learning systems}, vol.~27,
  no.~5, pp. 978--992, 2015.

\bibitem{yuan2017multi}
S.~Yuan, G.~Deng, Q.~Feng, P.~Zheng, and T.~Song, ``Multi-objective
  evolutionary algorithm based on decomposition for energy-aware scheduling in
  heterogeneous computing systems.'' \emph{J. UCS}, vol.~23, no.~7, pp.
  636--651, 2017.

\bibitem{olin2019stochasticity}
W.~Olin-Ammentorp, K.~Beckmann, C.~D. Schuman, J.~S. Plank, and N.~C. Cady,
  ``Stochasticity and robustness in spiking neural networks,'' \emph{arXiv
  preprint arXiv:1906.02796}, 2019.

\bibitem{schuman2014neuroscience}
C.~D. Schuman, J.~D. Birdwell, and M.~Dean, ``Neuroscience-inspired inspired
  dynamic architectures,'' in \emph{Proceedings of the 2014 Biomedical Sciences
  and Engineering Conference}.\hskip 1em plus 0.5em minus 0.4em\relax IEEE,
  2014, pp. 1--4.

\bibitem{elliott2016exploiting}
J.~Elliott, M.~Hoemmen, and F.~Mueller, ``Exploiting data representation for
  fault tolerance,'' \emph{Journal of computational science}, vol.~14, pp.
  51--60, 2016.

\end{thebibliography}

\end{document}